<u>Short Paper</u>

# Two-Layer Retrieval-Augmented Generation Framework for Low-Resource Medical Question Answering Using Reddit Data: Proof-of-Concept Study

Sudeshna Das[1*], PhD; Yao Ge[1*], PhD; Yuting Guo[1], MS; Swati Rajwal[2], MTech; JaMor Hairston[1], MS; Jeanne Powell[1], PhD; Drew Walker[1], PhD; Snigdha Peddireddy[3], MPH; Sahithi Lakamana[1], MS; Selen Bozkurt[1], PhD; Matthew Reyna[1], PhD; Reza Sameni[1,4], PhD; Yunyu Xiao[5], PhD; Sangmi Kim[6], PhD; Rasheeta Chandler[6], PhD; Natalie Hernandez[7], PhD; Danielle Mowery[8], PhD; Rachel Wightman[9], MD; Jennifer Love[10], MD; Anthony Spadaro[11], MD; Jeanmarie Perrone[12], MD; Abeed Sarker[1,4], PhD

[1]Department of Biomedical Informatics, School of Medicine, Emory University, Atlanta, GA, United States

[2]Department of Computer Science and Informatics, Emory University, Atlanta, GA, United States

[3]Department of Behavioral, Social & Health Education Sciences, Rollins School of Public Health, Emory University, Atlanta, GA, United States

[4]Department of Biomedical Engineering, Georgia Institute of Technology and Emory University, Atlanta, GA, United States

[5]Department of Population Health Sciences, Weill Cornell Medicine, New York, NY, United States

[6]Nell Hodgson Woodruff School of Nursing, Emory University, Atlanta, GA, United States

[7]Center for Maternal Health Equity, Morehouse School of Medicine, Atlanta, GA, United States

[8]Department of Biostatistics, Epidemiology and Informatics, Perelman School of Medicine, University of Pennsylvania, Philadelphia, PA, United States

[9]Department of Emergency Medicine, Warren Alpert Medical School of Brown University, Providence, RI, United States

[10]Department of Emergency Medicine, Icahn School of Medicine at Mount Sinai, New York, NY, United States

[11]Department of Emergency Medicine, Rutgers New Jersey Medical School, Newark, NJ, United States

[12]Department of Emergency Medicine, Perelman School of Medicine at the University of Pennsylvania, Philadelphia, PA, United States

[*]these authors contributed equally

**Corresponding Author:**
Sudeshna Das, PhD
Department of Biomedical Informatics
School of Medicine
Emory University
101 Woodruff Circle
Atlanta, GA, 30322
United States
Phone: 1 4047270229
Email: <u>sudeshna.das@emory.edu</u>

## Abstract

**Background:**   The increasing use of social media to share lived and living experiences of substance use presents a unique opportunity to obtain information on side effects, use patterns, and opinions on novel psychoactive substances. However, due to the large volume of data, obtaining useful insights through natural language processing technologies such as large language models is challenging.

**Objective:**   This paper aims to develop a retrieval-augmented generation (RAG) architecture for medical question answering pertaining to clinicians' queries on emerging issues associated with health-related topics, using user-generated medical information on social media.

**Methods:**   We proposed a two-layer RAG framework for query-focused answer generation and evaluated a proof of concept for the framework in the context of query-focused summary generation from social media forums, focusing on emerging drug-related information. Our modular framework generates individual summaries followed by an aggregated summary to answer medical queries from large amounts of user-generated social media data in an efficient manner. We compared the performance of a quantized large language model (Nous-Hermes-2-7B-DPO), deployable in low-resource settings, with GPT-4. For this proof-of-concept study, we used user-generated data from Reddit to answer clinicians' questions on the use of xylazine and ketamine.

**Results:** Our framework achieves comparable median scores in terms of relevance, length, hallucination, coverage, and coherence when evaluated using GPT-4 and Nous-Hermes-2-7B-DPO, evaluated for 20 queries with 76 samples. There was no statistically significant difference between GPT-4 and Nous-Hermes-2-7B-DPO for coverage (Mann-Whitney $U=733.0$; $n_1=37$; $n_2=39$; $P=.89$ two-tailed), coherence ($U=670.0$; $n_1=37$; $n_2=39$; $P=.49$ two-tailed), relevance ($U=662.0$; $n_1=37$; $n_2=39$; $P=.15$ two-tailed), length ($U=672.0$; $n_1=37$; $n_2=39$; $P=.55$ two-tailed), and hallucination ($U=859.0$; $n_1=37$; $n_2=39$; $P=.01$ two-tailed). A statistically significant difference was noted for the Coleman-Liau Index ($U=307.5$; $n_1=20$; $n_2=16$; $P<.001$ two-tailed).

**Conclusions:** Our RAG framework can effectively answer medical questions about targeted topics and can be deployed in resource-constrained settings.

## Introduction

Large language models (LLMs) present opportunities for solving complex biomedical natural language processing problems, such as medical question answering (MQA). However, operational challenges (eg, high computational resource requirements) hinder their real-life deployment and use. Another issue with LLM-generated text for MQA is "hallucination": generated text that is plausible-sounding but nonsensical or incorrect [1]. Chain-of-thought prompting [2], self-reflection [1], and retrieval-augmented generation (RAG) are forerunners in mitigating hallucination. RAG also aids in constraining generated texts and improves in-context learning [3]. LLMs in RAG frameworks have been used in the biomedical domain owing to the need for timely, accurate, and transparent responses [4]. As LLMs become increasingly integrated into clinical practice [5], it is important to ensure their operability in low-resource settings [6] while generating accurate and coherent texts.

We present a proof-of-concept study for a two-layer RAG framework for MQA that ingests user-generated medical information from Reddit. We used smaller, quantized, open-source LLMs that can run on personal computers without specialized hardware, allowing our framework to be used in low-resource settings, thus ensuring equitable access to timely medical information.

## Methods

### Study Design

We evaluated our proof-of-concept study in a setting where copious amounts of data are available for a topic but gathering insights and answering questions require substantial manual work—the topic of emerging drugs from Reddit. Reddit has ~52 million daily active users, is commonly used to study emerging medical themes [7], and features numerous discussions on the nonmedical uses of substances. Recently, Reddit data have been leveraged to study novel psychoactive substances since such information is not typically available elsewhere. We chose two substances that have gained attention recently—xylazine (because of its increasing impact and association with the US opioid crisis) and ketamine (because of its recent popularity as a treatment for depression). We collected all available data (~2.5 billion posts) from Reddit until December 31, 2023, and extracted all posts mentioning xylazine (n=177,684) and ketamine (n=7699) for our retrieval engine. Based on clinician-driven interests, we formulated 20 queries (Table 1).

**Table 1.** Queries used for evaluating the framework.

| Query ID | Query |
|---|---|
| 1 | What are the side effects of xylazine? |
| 2 | What does xylazine do to the skin? |
| 3 | How does xylazine impact rehab? |
| 4 | What is xylazine withdrawal like? |
| 5 | What drugs contain xylazine? |
| 6 | What treatments work for xylazine? |
| 7 | What drugs are mixed or cut with xylazine? |
| 8 | What areas of the United States are impacted by xylazine? |
| 9 | How is xylazine different from pure heroin? |
| 10 | What is the general sentiment associated with xylazine? |
| 11 | Does narcan or naloxone work for xylazine overdose? |
| 12 | What are the side effects of ketamine? |
| 13 | What is ketamine withdrawal like? |
| 14 | What are k cramps like? |
| 15 | How do the users describe k hole? |
| 16 | Does ketamine work for depression? |
| 17 | What drugs are ketamine coused with recreationally? |
| 18 | Is ketamine effective for the treatment of suicidal behavior? |
| 19 | How can you treat ketamine addiction? |
| 20 | Does ketamine use cause cramps? |

## System Architecture

As depicted in Figure 1, the user submits a query to be parsed by the information retrieval engine, which returns a ranked list of documents. The top n documents are chosen to be sources for answer generation. In the first layer, the LLM is provided with (1) a query, (2) text from the retrieved documents, and (3) a prompt that embeds the text and instructs the LLM to summarize it (Multimedia Appendix 1). Since the prompt context window is finite, feeding the LLM all the retrieved text for answer generation is typically impossible. Even single documents can be too long. Thus, the framework allows for the specification of segment lengths for the retrieved text in each iteration, ensuring the framework is applied to relatively small LLMs with shorter context lengths. The first layer generates short, query-focused summaries (Figure 1). The LLM states if the retrieved text segment does not contain an answer to the question. Examples of this first-layer summarization are provided in Multimedia Appendix 2.

The second layer takes as input the original query and individual short summaries embedded within a second prompt (Multimedia Appendix 1) while ignoring summaries that the LLM states did not contain the answer. Figure 1 depicts an example of the final, synthesized summary.

We used the 8-bit quantized model Nous-Hermes-2-7B-DPO as our LLM, which is tuned on 1,000,000 high-quality instructions [8]. To test the performance of the proposed framework with larger models, we also performed an evaluation using GPT-4 [9].

**Figure 1.** Overview of the two-layer RAG framework. The first layer generates individual summaries based on retrieved posts relevant to the original query. The second layer generates the final summary based on the individual summaries generated in the first layer. LLM: large language model; RAG: retrieval-augmented generation.

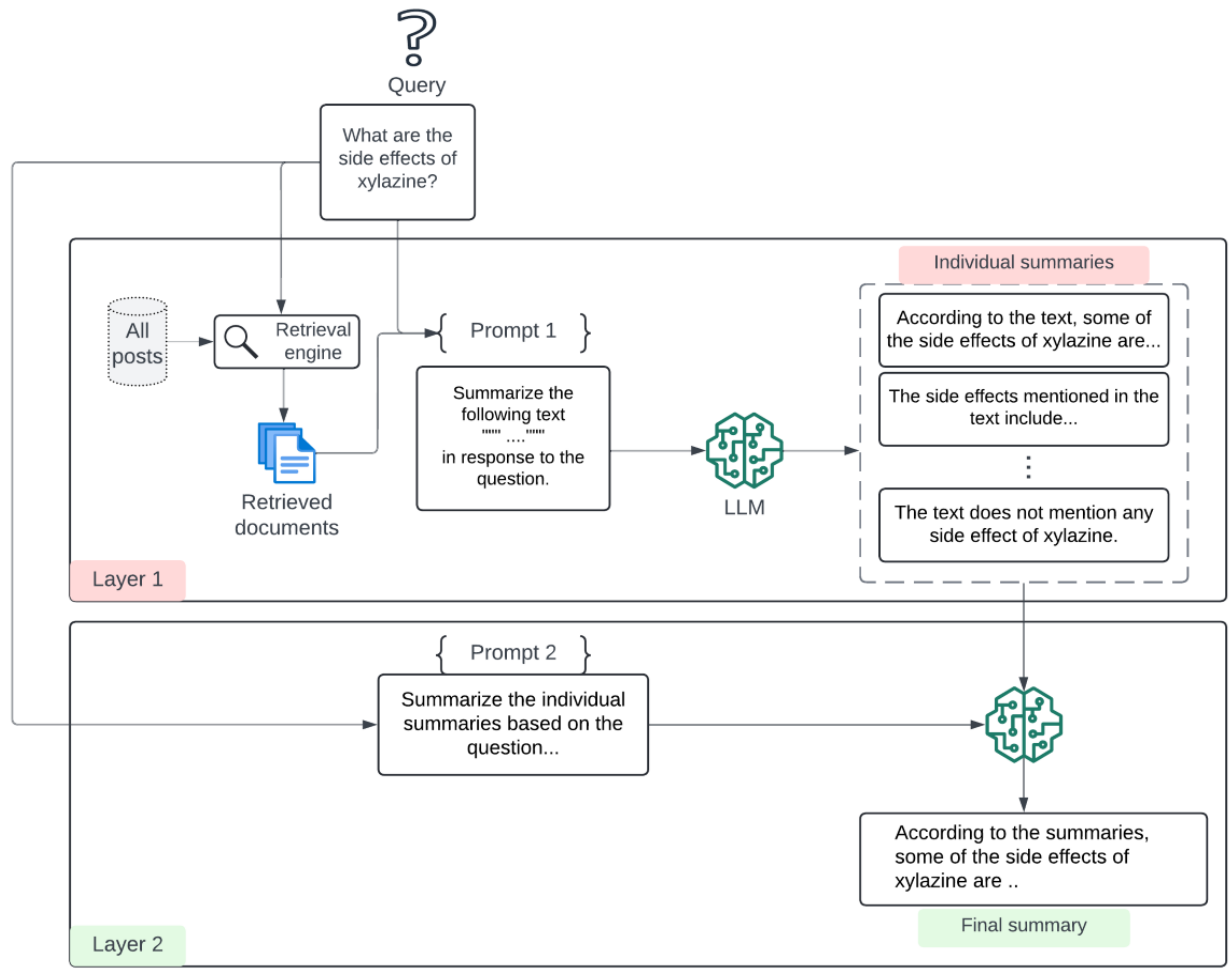

## RAG Architecture

Since the retrieval aspect is not our primary focus, we used a simple keyword-based approach using the default search settings provided by our information retrieval package "Whoosh," which uses Okapi BM25F ranking [10]. The top 50 retrieved documents were chosen for generating the first-layer summaries. This number may be adjusted without changes to the architecture. The number of text segments is typically higher when posts do not fit within the context window of the LLM after being embedded within the prompt.

## Evaluation

Our evaluation focused on the architecture's summary generation quality, rather than retrieval performance. Commonly used automatic summary evaluation methods, such as Recall-Oriented Understudy for Gisting Evaluation (ROUGE) [11] and bilingual evaluation understudy (BLEU) [12], primarily focus on text overlap between generated and gold-standard summaries. In the absence of gold-standard summaries, subject matter experts manually and qualitatively evaluated the important nuances of generative summaries, which is impossible with ROUGE or BLEU. We used Likert-scale evaluations (Table 2). Each query–individual summary–final summary triplet was evaluated by ≥2 evaluators (at least a master's degree in medicine, public health, informatics, or allied fields). Overall, 21 experts generated 76 evaluations for 20 unique queries.

We also assessed the readability of the final summaries using the Coleman-Liau Readability Index (CLI) [13], which approximates the US grade level required to comprehend text.

We performed nonparametric tests for proportions (Mann-Whitney $U$ test) with the null hypothesis ($H_0$: "The two populations are equal") to determine if the scores assigned to answers generated by GPT-4 and Nous-Hermes-2-7B-DPO vary significantly. All tests were performed using the *SciPy* package [14]. The null hypothesis ($H_0$) was rejected if $P < .05$ (two-tailed).

**Table 2.** Evaluation criteria and scales presented to annotators.

| Criteria | Question | Evaluation scale |
|---|---|---|
| Coverage | Does the final summary accurately represent the information present in the original text? | • 5: Yes; the final summary covers all the important information present in the original text.<br>• 4: Mostly; the final summary covers most, but not all of the important information.<br>• 3: Somewhat; the final summary covers some of the important information, but also misses some of them.<br>• 2: Not really; the final summary misses most of the important information.<br>• 1: No; the final summary does not cover any of the important information present in the original text. |
| Coherence | Is the final summary coherent? | • 5: Yes; the final summary is easy to read and understand.<br>• 4: Mostly; the final summary is readable, but not straightforward to understand.<br>• 3: Somewhat; the final summary is readable but confusing.<br>• 2: Not really; the final summary has some grammatical errors or nonsequiturs.<br>• 1: No; the final summary is unintelligible or incomprehensible. |
| Relevance | Does the final summary answer the original question? | • 3: Yes; the summary answers the original question.<br>• 2: Partially; the summary answers the original question, but not fully.<br>• 1: No; the summary does not answer the original question. |
| Length | Is the length of the final summary appropriate? | • 3: Yes; the summary is appropriate in length.<br>• 2: Somewhat; the summary could be shorter or longer.<br>• 1: No; the summary is long-winded or too short. |
| Hallucination | Does the summary contain information not present in the original text? | • 0: No; the summary does not contain information not present in the original text.<br>• 1: Yes; the summary contains information not present in the original text. |

## Ethical Considerations

This study was deemed to be exempt from review per the Emory University Institutional Review Board's guidelines. The data used in this study are anonymous by default. We ensured that self-disclosed, personally identifiable information is not used by only reporting aggregated data. We removed posts that were deleted by the user.

## Results

We conducted extensive expert evaluations of the generated answers for coverage, coherence, relevance, length, and hallucination (Figure 2). Annotators were not made aware of which LLM was used to generate the summaries for fair evaluation. On a 5-point Likert scale, median coverage scores were 5 (IQR 4-5) for both; the distributions did not differ significantly ($U=733.0$; $n_1=37$; $n_2=39$; $P=.89$). Median coherence scores were 5 (IQR 5-5 for GPT-4; IQR 4-5 for Nous-Hermes-2-7B-DPO) for both; they did not differ significantly ($U=670.0$; $n_1=37$; $n_2=39$; $P=.49$).

On a 3-point Likert scale, the median relevance scores were 3 (IQR 3-3) for both; they did not differ significantly ($U=662.0$; $n_1=37$; $n_2=39$; $P=.15$). Median length scores were 3 (IQR 2-3) for both; they did not differ significantly ($U=672.0$; $n_1=37$; $n_2=39$; $P=.55$). On a binary Likert scale, median hallucination scores were 0 (IQR 0-0) for both; they did not differ significantly ($U=859.0$; $n_1=37$; $n_2=39$; $P=.10$). The median CLIs were 16.635 (IQR 13.860-17.675) for GPT-4 and 12.125 (IQR 11.02-13.98) for Nous-Hermes-2-7B-DPO; there was a statistically significant difference ($U=307.5$; $n_1=20$; $n_2=16$; $P<.001$).

Median token counts for queries posed to GPT-4 and Nous-Hermes-2-7B-DPO were 5 (IQR 5-7) and 7 (IQR 5-8), respectively; there was no significant difference ($U=165.0$; $n_1=20$; $n_2=16$; $P=.66$). Median lengths of responses generated by GPT-4 and Nous-Hermes-2-7B-DPO were 1118 (IQR 709-2986) and 441 (IQR 231-695) for the combined individual summaries and 141.5 (IQR 115-159) and 61 (IQR 28-87) for the final summaries, respectively. Both were significantly different ($U=300.0$; $n_1=20$; $n_2=16$; $P<.001$, and $U=145.5$; $n_1=20$; $n_2=16$; $P<.001$).

**Figure 2.** Box plots illustrating the distribution of scores for the evaluation criteria used. (A) Coverage on a 5-point Likert scale. (B) Coherence on a 5-point Likert scale. (C) Relevance on a 3-point Likert scale. (D) Length on a 3-point Likert scale. (E) Hallucination on a binary scale. (F) Values for the Coleman-Liau Index. (G) Token counts for Questions. (H) Token counts for combined individual summaries. (I) Token counts for the final summary.

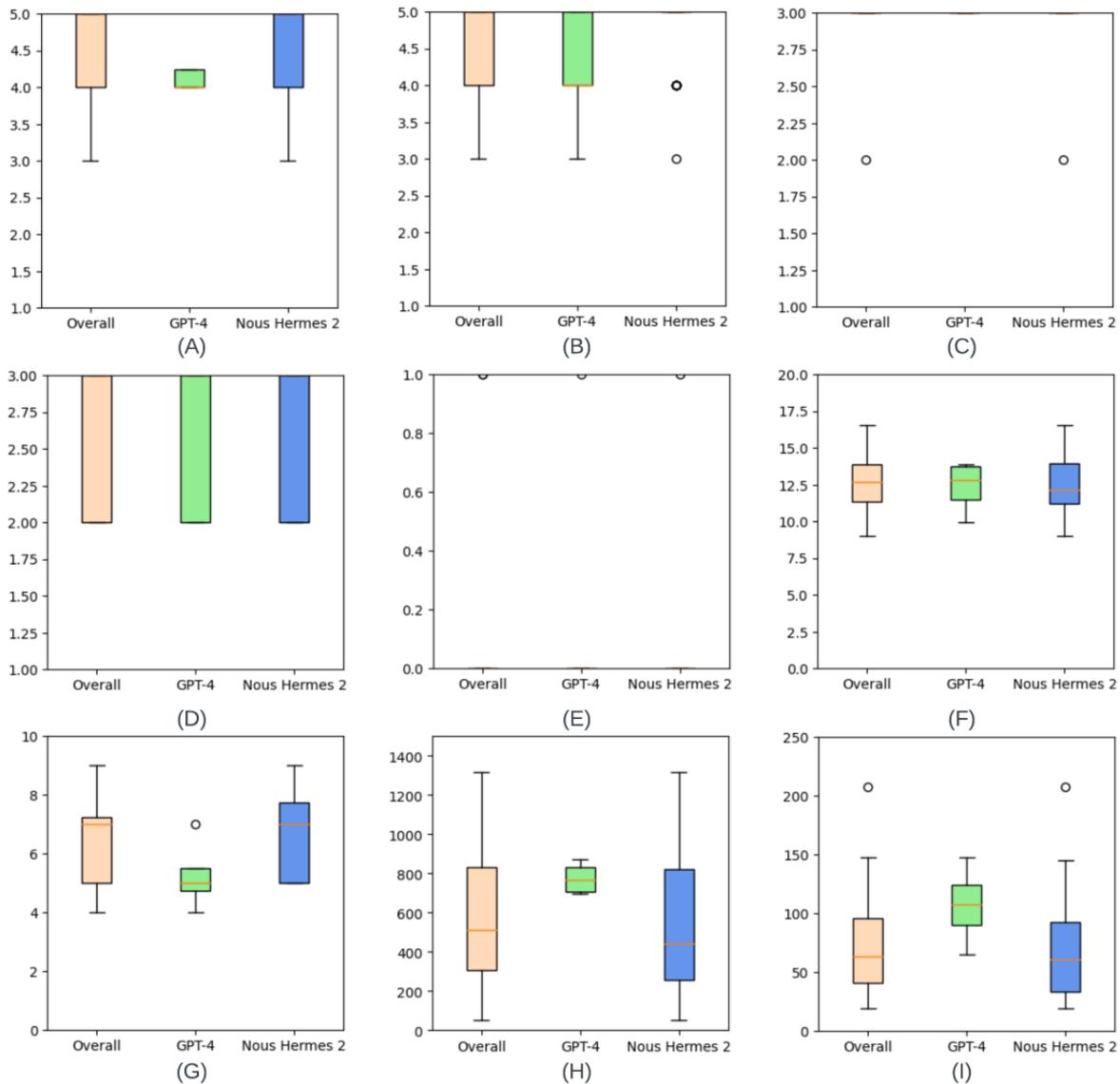

## Discussion

### Principal Findings

This study presents a novel two-layer RAG framework for MQA that uses user-generated content from Reddit. Our findings demonstrate that the framework effectively synthesizes accurate and contextually relevant answers even in low-resource settings, aligning with our goal to create an accessible, computationally lightweight tool. Focusing on small, quantized, open-source LLMs ensures equitable access to valuable insights about emerging trends, potential side effects, and general perception of substances, as reflected in Reddit posts.

The modular structure of the framework enables good performance without requiring specialized hardware, which is critical in low-resource environments. This modularity also supports using different retrieval engines or LLMs, providing flexibility to adapt to various use cases. The system's ability to answer nuanced queries (eg "What are k cramps like?"—which would require extensive manual curation) illustrates its potential for real-world applications. The framework's ability to specify temporal ranges in queries allows it to track trends over time, offering opportunities for longitudinal studies and misinformation detection.

Unlike previous work [15], where segments of text are generated chronologically, we performed segmentation at the post level without accounting for chronology. Compared with existing literature [4,16], this study underscores the potential of smaller models for tasks requiring domain-specific, contextually accurate outputs. Prior work often focuses on high-resource settings [17,18] with robust computational infrastructure, leaving gaps in applicability for low-resource environments. We fill this gap by showing that reliable performance can be achieved with computationally efficient architectures, expanding the reach of artificial intelligence tools to underresourced regions. Although smaller LLMs have been used [19], summarization from large volumes of text with aggregated information has not been addressed before.

## Limitations

Despite its promising features, the framework has limitations. It relies on the accuracy and representativeness of the Reddit data it ingests. Reddit posts may include biases, inaccuracies, or misinformation that could influence the system's output.

While faithfully summarizing misinformation is valuable for transparency, users need to exercise caution in interpreting the results. Additionally, we evaluated the framework using a small set of queries pertaining to substance use; further validation is necessary to assess performance across diverse medical domains.

## Conclusions

This study demonstrates that a modular, lightweight RAG framework can effectively address complex MQA using social media data in low-resource settings. By enabling clinicians to rapidly extract insights about substance use trends and potential side effects from Reddit posts, the framework holds significant potential for improving public health.

## Acknowledgments

Research reported in this publication was supported by the National Institute on Drug Abuse of the National Institutes of Health (NIH) under award R01DA057599. The content is solely the responsibility of the authors and does not necessarily represent the official views of the NIH.

## Data Availability

All data used in this study were publicly available from Reddit at the time of data collection. The second-level summaries are available in Multimedia Appendix 3. The original posts and social media posts analyzed during this study are available from the corresponding author upon reasonable request and the completion of a data use agreement.

## Authors' Contributions

SD and Y Ge led analysis, evaluation, visualization, and original draft preparation. Y Guo, SR, JH, J Powell, DW, SP, and SL contributed to the evaluation and manuscript preparation. SB, MR, RS, YX, SK, RC, NH, DM, RW, JL, A Spadaro, and J Perrone worked on the evaluation. A Sarker conceptualized the study, led model design and implementation, and supervised the project.

## Conflicts of Interest

None declared.

## Multimedia Appendix 1

Prompts used.
[DOCX File , 12 KB-Multimedia Appendix 1]

## Multimedia Appendix 2

Sample first-layer individual summaries.
[DOCX File , 14 KB-Multimedia Appendix 2]

## Multimedia Appendix 3

Final summaries generated by the framework for each of the 20 queries used for evaluation.
[DOCX File , 30 KB-Multimedia Appendix 3]

## Abbreviations

**BLEU:** Bilingual Evaluation Understudy
**CLI:** Coleman-Liau Index
**LLM:** large language model
**MQA:** medical question answering
**RAG:** retrieval-augmented generation
**ROUGE:** Recall-Oriented Understudy for Gisting Evaluation